\documentclass[conference]{IEEEtran}
\IEEEoverridecommandlockouts
\usepackage{cite}
\usepackage{amsmath,amssymb,amsfonts}
\usepackage{graphicx}
\usepackage{textcomp}
\usepackage{xcolor}
\usepackage{microtype}
\usepackage{graphicx}
\usepackage{subfig}
\usepackage{booktabs}
\usepackage{multirow}
\usepackage{amsmath}
\usepackage{algorithm}
\usepackage{algpseudocode}
\usepackage{etoolbox}
\usepackage{amsmath,amsfonts}
\usepackage{graphicx}
\usepackage{xcolor}
\usepackage{hyperref}
\def\BibTeX{{\rm B\kern-.05em{\sc i\kern-.025em b}\kern-.08em
    T\kern-.1667em\lower.7ex\hbox{E}\kern-.125emX}}

\begin{document}

\author{\IEEEauthorblockN{Youshao Xiao\textsuperscript{\textdagger} \thanks{{\textdagger} Equal contributions},
Zhenglei Zhou\textsuperscript{\textdagger},
Fagui Mao\textsuperscript{\textdagger},
Weichang Wu}
\IEEEauthorblockN{
Shangchun Zhao,
Lin Ju,
Lei Liang, 
Xiaolu Zhang,
Jun Zhou\textsuperscript{*} \thanks{* Corresponding authors}
}
\IEEEauthorblockA{
\textit{Ant Group, Hangzhou, China}\\
\{youshao.xys, zhouzhenglei.zzl, maofagui.mf, jiuyue.wwc, shangchun.zsc, julin.jl\}@antgroup.com \\
\{leywar.liang, yueyin.zxl, jun.zhoujun\}@antgroup.com
}}

\title{An Adaptive Placement and Parallelism Framework for Accelerating RLHF Training}

\maketitle

\begin{abstract}
Recently, ChatGPT or InstructGPT like large language models (LLM) has made a significant impact in the AI world. Many works have attempted to reproduce the complex InstructGPT’s training pipeline, namely Reinforcement Learning with Human Feedback (RLHF). However, the mainstream distributed RLHF training methods typically adopt a fixed model placement strategy, referred to as the Co-located strategy. This strategy treats all four interdependent models involved in RLHF as a single entity, distributing them across all devices and applying parallelism techniques designed for a single model, regardless of the workload heterogeneity inherent to each model. As a result, this strategy exacerbates the generation bottlenecks in the RLHF training and degrades the overall training efficiency. To address these issues, we propose a flexible model placement framework that offers two general and agile model placement strategies. The Interleaving strategy helps reduce memory redundancy and communication costs of RLHF training by placing models without dependencies on exclusive devices with careful orchestration. On the other hand, the Disaggregated strategy improves the throughput of model training by separating the training and inference runtime of the RLHF pipeline with additional shadow models. Furthermore, our framework provides a simple user interface and guidelines to easily and flexibly configure these strategies in various training scenarios. Our experiments have shown that our strategy can achieve notable improvements up to 11$\times$, compared to the current state-of-the-art (SOTA) approaches. The results highlight the effectiveness and adaptability of our methods in accelerating the training of distributed RLHF.
\end{abstract}


\section{Introduction}
Large Language Models (LLMs) have demonstrated remarkable capabilities across various natural language processing tasks~\cite{brown2020language,huang2023mvp,yang2024qwen2,chowdhery2023palm, touvron2023llama}. However, it is crucial to fine-tune LLM models to align them with human preferences. In the absence of proper alignment with human feedback, LLMs may exhibit behaviors that deviate from expectations. These behaviors include generating fabricated information, pursuing inaccurate objectives, and producing harmful, misleading, and biased expressions~\cite{ouyang2022training,kenton2021alignment, glaese2022improving,weidinger2022taxonomy}). To tackle this issue, researchers have proposed a series of approaches aimed at ensuring that LLMs adhere to ethical considerations and societal norms.
One of the most successful approaches in addressing the alignment issue is Reinforcement Learning from Human Feedback (RLHF)~\cite{stiennon2020learning,achiam2023gpt,bai2022training,christiano2023deep}. Particularly, the pipeline proposed in InstructGPT ~\cite{ouyang2022training}, a sibling model of ChatGPT, has gained significant attention. RLHF combines human alignment with Reinforcement Learning (RL)~\cite{sutton2018reinforcement,hu2021enumeration,mnih2013playing}, a decision-making approach that allows AI agents to learn from interactions with the environment, receiving feedback in the form of rewards. This integration of RL with Human Feedback (HF) enables researchers to enhance the capabilities of LLMs and improve their alignment with human values. 

\begin{figure}[ht]
  \centering
  \includegraphics[width=1.0\linewidth]{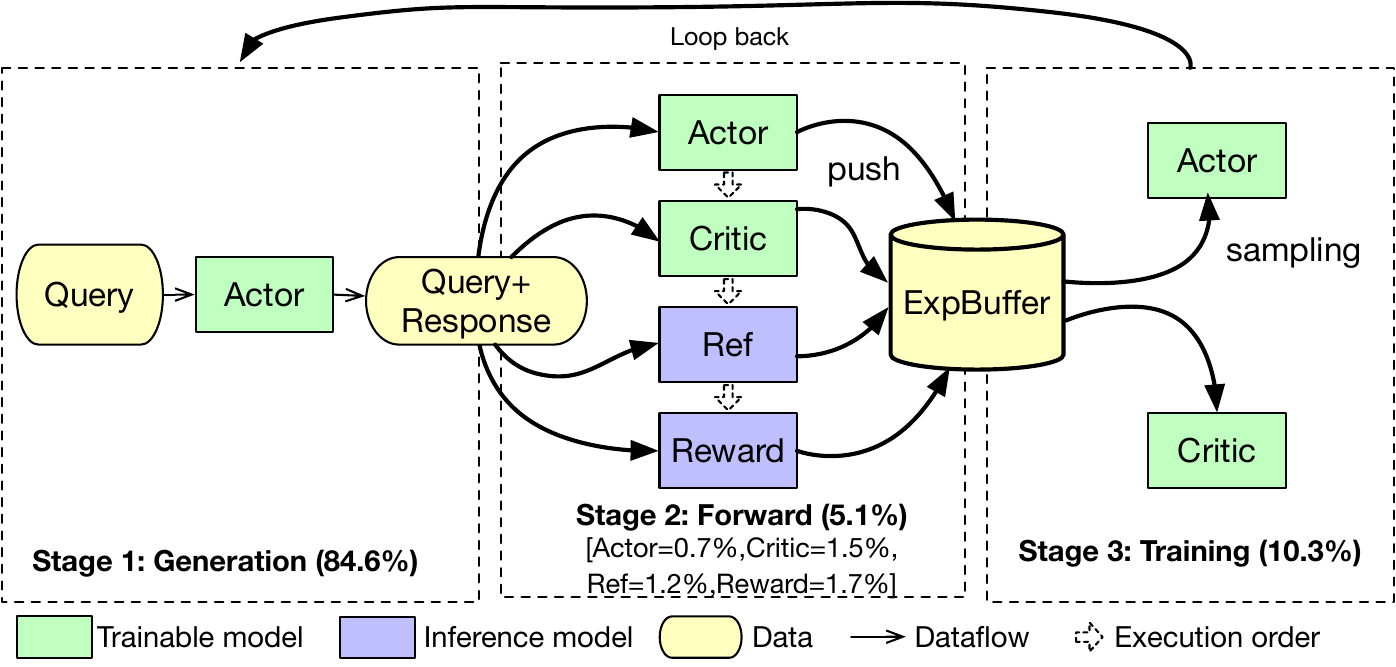}
  \caption{Workflow of RLHF and Percentage (in \%) of Each Stage Duration.}
  \label{fig:percentage}
\end{figure}

The predominant RL algorithm to finetune LLM is Proximal Policy Optimization (PPO) ~\cite{schulman2017proximal,ouyang2022training,peng2023graphrare} and is known as ``Step-3: Training RL Policy" in InstructGPT. This process leverages an Actor-Critic architecture~\cite{mnih2016asynchronous,sutton2018reinforcement,ouyang2022training}, comprising four \textbf{interconnected} models: a Reference (Ref) model, a Reward model, an Actor model, and a Critic model. The pipeline is segmented into three sequential stages—generation, forward, and training—as depicted in Fig. \ref{fig:percentage}. Note that Actor and Critic models are \textbf{trainable models}, which are leveraged for \emph{both training and inference runtime} while Ref and Reward models are \textbf{inference models} utilized \emph{solely for
inference}. Specifically, in the generation stage, the Actor model generates the response from the query. In the subsequent forward stage, given the query and response, all four models follow a \textbf{sequential execution order} to produce intermediate output results such as logits or values. For example, the Actor, Reference, Reward, and Critic models conduct the forward stage one after the other. These results are stored in the experience replay buffer. During the training stage, the Actor and Critic models are trained using batch data sampled from the buffer. This whole procedure repeats over a number of steps until it converges. It is important to note that the Actor and Critic models utilize the same underlying model but under a \textbf{mixed usage of training and inference runtime} across these three stages.

Across the training of RLHF, the size of individual LM parameters can exhibit significant variations, e.g. spanning from 7 billion (7B) to 65 billion (65B) parameters in models ~\cite{touvron2023llama,dubey2024llama}. The presence of LMs in the RLHF pipeline amplifies memory usage quadratically, which underscores the critical need for efficient distributed training methods. However, distributed training techniques, such as data parallelism or model parallelism~\cite{xiao2023g,rajbhandari2020zero,shoeybi2019megatron}, are primarily designed for training a single model, not for efficiently managing multiple interdependent models as seen in RLHF. This necessitates the development of innovative model placement strategies. 

Existing works~\cite{yao2023deepspeed,havrilla2023trlx,li2023colossal} typically employ a straightforward model placement strategy, namely \textbf{Co-located strategy}. \emph{This strategy treats all involved models as a single entity and applies standard parallelism techniques designed for individual models.} While this strategy aligns well with the sequential execution order as the single-machine PPO algorithm, it brings several limitations and opportunities for optimization. 1) This approach results in reduced memory available for a single model on each device. Therefore, we have to add up more devices and parallelism techniques to accommodate all four models, with increased memory redundancy and communication costs ~\cite{rajbhandari2020zero, wu2023rethinking,smith2022using}. 2) Different scales of model sizes require fine-grained model placement strategies. For instance, 7B models could be held by fewer devices compared with 65B models, which creates the opportunity for optimization. 3) The workloads during training and inference runtime inherently differ, necessitating distinct optimization techniques~\cite{wu2023rethinking,kwon2023efficient}. Specifically, in the training stage, it is advisable to shard the optimizer state, gradients, and parameters sequentially for better efficiency. However, only the parameters need to be shared during the inference stage, as gradients and optimizer state are not required during inference~\cite{aminabadi2022deepspeed}.

Therefore we propose \textbf{FlexRLHF}, a flexible Reinforcement Learning from Human Feedback (RLHF) framework, that significantly enhances the training efficiency in distributed RLHF training while ensuring no compromise on model performance. Firstly, 
our framework introduces two innovative model placement strategies to improve training efficiency in various RLHF training scenarios. Secondly, we design a FlexRLHF Execution Engine abstraction that provides a user-friendly interface to modeling experts, shielding them from the intricate details of model placement and parallelism strategies. Thirdly, we provide a guideline on how to efficiently configure these distributed strategies in different training scenarios.

Our work makes four main contributions:

i) \textbf{Flexible Model Placement:} Our FlexRLHF framework introduces two flexible model placement strategies. Our Interleaving strategy places models without dependencies on exclusive devices and our Disaggregated strategy decouples the training and inference runtime of the RLHF pipeline, significantly improving the training efficiency.

ii) \textcolor{ black}{\textbf{Heterogeneity Support:} Our framework accommodates diverse runtime environments, including different distributed training engines (e.g., DeepSpeed and Megatron) and inference engines. It also supports GPUs with varied capabilities and costs.}

iii) \textbf{Superior Performance:} Our extensive experiments confirm that our method outperforms the state-of-the-art up to $11\times$ in terms of throughput. 

iv) \textbf{Ease of Use:} Our framework features an Execution Engine with guidelines to support various model placement strategies and acceleration techniques with minimal or no code changes. 

\section{Background}\label{sec:background}

\subsection{Data and Model parallelism}\label{sec:background_dp}
According to different parallel objects, distributed training technology for parallelizing single language models can be divided into data parallelism and model parallelism.

Data Parallelism (DP) involves dividing the input data into multiple partitions and assigning them to different devices for parallel processing. In traditional DP training methods like AllReduce~\cite{gibiansky2017bringing, sergeev2018horovod} or Distributed Data Parallel (DDP)~\cite{li13pytorch,paszke2019pytorch}, each device holds a complete replica of the model states and performs the \textit{AllReduce} primitive to synchronize the model states across devices. ZeRO (Zero Redundancy Optimizer)~\cite{rajbhandari2020zero} is a technique that addresses memory redundancy of the model state replicas held on each device, which dominates the DP-based LLM training. However, the increased level of ZeRO parallelism helps reduce memory redundancy but comes with growing communication costs~\cite{rajbhandari2020zero,wu2023rethinking}.

Model Parallelism, including Tensor and Pipeline Parallelism, involves distributing model parameters across multiple devices within or across instances~\cite{shoeybi2019megatron}. Tensor Parallelism (TP)~\cite{shazeer2018mesh,shoeybi2019megatron} achieves this by vertically splitting the model, sharding tensors across multiple devices, and computing through distributed matrix multiplication. However, due to the frequent global communication, Tensor Parallelism is more efficient within nodes with a high GPU interconnect bandwidth. On the other hand, Pipeline Parallelism (PP)~\cite{huang2019gpipe,harlap2018pipedream} adopts a horizontal model splitting approach, assigning different layers to different devices. It usually uses micro-batching to hide the bubbles which is the idle duration of the device in the pipeline. 

\subsection{Heterogeneous Network}\label{sec:background_network}
Clusters with heterogeneous networks are common in cloud computing environments~\cite{jia2022whale}, where the bandwidth between intra-node and inter-node is unbalanced. Typically, the inter-node bandwidth can be from 3x to 24x times slower than the intra-node bandwidth~\cite{MiCS22,wang2023zero++}. This limitation becomes even more pronounced in heterogeneous GPU clusters, such as a mixture of A100 and V100, where different types of GPU devices can only be connected using Ethernet. The significant gap in the network between intra- and inter-node greatly hampers the communication efficiency of Tensor Parallelism and Data Parallelism, especially at higher levels of ZeRO~\cite{wu2023rethinking,rajbhandari2020zero}. Moreover, as the size of the cluster scales out, the communication overhead caused by heterogeneous networks grows larger, which downgrades the training efficiency.

\section{Related Works and Motivation}
\subsection{Related Works}
Since the emergence of ChatGPT and InstructGPT, several distributed frameworks have been proposed to support the parallelization of complex InstructGPT-like RLHF training pipelines ~\cite{yao2023deepspeed,vonwerra2022trl,li2023colossal,havrilla2023trlx}.  From a modeling aspect, these works primarily consider two RLHF model structures: \textbf{AC-Share}~\cite{havrilla2023trlx} and \textbf{AC-NonShare}~\cite{yao2023deepspeed}, \textit{depending on whether the Actor and Critic model share the parameters or not}. For example, in the \textbf{AC-Share structure}, \textit{the Actor and Critic models share parameters by adding an extra linear layer to differentiate between them, and vice versa}. This reduces the memory and resource requirements for RLHF training, but it may potentially lead to a loss in model performance.

To parallelize the training of multiple interdependent models in the RLHF pipeline, all of these works adopt a fixed model placement strategy or its variant, referred to as the \textbf{Co-located strategy}~\cite{yao2023deepspeed,vonwerra2022trl,li2023colossal,havrilla2023trlx}. \emph{The Co-located strategy straightforwardly takes four models as a whole and treats them equally.} For instance, DeepSpeed-Chat~\cite{yao2023deepspeed}, one of the most popular RLHF training frameworks using the AC-NonShare structure, places all four models of RLHF pipeline on each device and applies data and model parallelism techniques like ZeRO to parallelize the training. trlX~\cite{havrilla2023trlx}, another well-known work utilizing the AC-Share structure, employs a variant of the Co-located placement strategy. It places the Reward model on a fully occupied device to leverage its computational capabilities exclusively while deploying the Actor, Critic (shared with the Actor model), and Ref model on other devices using the Co-located strategy.

However, there is an apparent issue in the trlX framework. By placing the Reward model on a single GPU device, it underutilizes both memory and computational power. If the Reward model size is too small, GPU memory is wasted, while a Reward model that is too large may lead to out-of-memory issues. Additionally, the device exclusively allocated to the Reward model will experience significant idle time within a single step. Furthermore, extra communication is required to gather the generated results using \textit{AllGather} from the Actor model, and \textit{Scatter} the prediction outputs back to the Actor model placed on other devices. These problems significantly degrade the overall efficiency.

\textcolor{ black}{In addition, Deepspeed-Chat further proposes a Hybrid Engine~\cite{yao2023deepspeed}. During the training phase, it utilizes ZeRO for efficient memory management, while switching to tensor parallelism for the generation phase. To reshard model states between training and inference runtime, it performs the \textit{AllGather} operation to gather the whole parameters into tensor parallel workers in each generation. Although it has avoided frequent \textit{AllGather} communication in each forward for one generation, the remaining one for each generation still introduces significant overhead.}

\subsection{Limitations in Existing Distributed RLHF Training}\label{sec:limitation}
Without loss of generality, we take the AC-Share structure as an example. Fig. \ref{fig:percentage} illustrates that the Actor Generation accounts for more than 85\% of the total duration in one step, while the Training stage only occupies 10\% and then other forward stages when using DeepSpeed-Chat. The generation stage significantly slows down the entire RLHF training. The bottleneck comes from two main aspects: sequential execution and the fixed model placement strategy.

\textbf{Sequential Execution.} Existing works typically follow a sequential execution order, similar to the single-machine RLHF algorithm as shown in Fig. \ref{fig:percentage}, where each stage or model is executed one after another. However, the generation stage is essentially more computation-intensive than other stages. The generation stage primarily involves a time-consuming auto-regressive generation using the Actor model, followed by four sequential forward phases in each model. In this sequential execution order, the intermediate results generated from the time-cost generation stage in the experience replay buffer are rapidly consumed by the following stages, namely the forward and training stages. However, the training model consumes typically eight times memory as much as the inference model. Consequently, the training models are mostly idle but occupy a large amount of memory.

\begin{figure*}[tb!]
  \centering
  \includegraphics[width=1\textwidth]{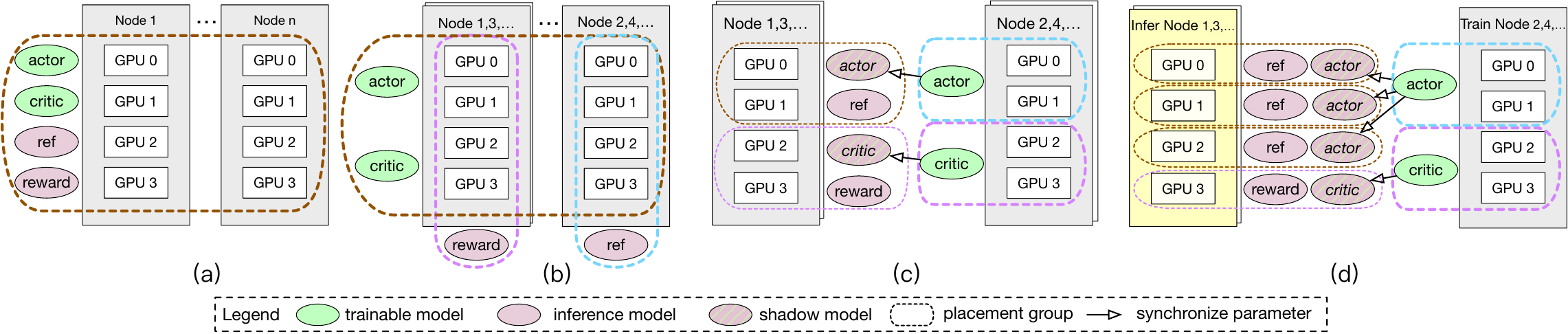}
  \caption{The architecture of Model Placement Strategies, where (a) represents the Co-located strategy, (b) represents the Interleaving strategy, (c) demonstrates the Disaggregated strategy used for homogeneous devices, where generation and training models are assigned to exclusive groups of devices, and (d) showcases the Disaggregated strategy used for heterogeneous devices, where inference models are allocated to dedicated groups of devices specialized for inference.}\label{fig:placement}
\end{figure*}

\textbf{Fixed Model Placement Strategy.} 
These works primarily rely on a fixed model placement strategy, namely the Co-located strategy, regardless of the size of the models used. It is easy to implement without additional intermediate data interaction and ensures that all devices keep working throughout all three stages but depends on sequential execution. \footnote{Someone may argue a simple placement strategy to remove the sequential execution order. It could exclusively place four models engaged in the Forward stage on separate groups of devices for parallelization, given non-data dependencies among these four models during the forward stage. Nonetheless, this strategy fails to consider the interdependencies between the models across three stages. For example, the Actor model will be also utilized in the generation and training stage. During the generation stage, only devices with Actor models work, leaving other devices idle. The Co-located Strategy is more efficient than this strategy, as it ensures all devices are actively engaged across all three stages. However, as discussed, it results in additional memory redundancy and costly inter-node communication, which downgrades the throughput.} However, this lack of adaptability does possess three limitations:

1) \emph{The Co-located strategy incurs memory redundancy and additional communication costs in RLHF training}.  It deploys all four models on every device, significantly reducing the available memory on each
device for a single model. This necessitates an increased level of ZeRO parallelism to accommodate all models but with extra memory redundancy and communication costs, as discussed in section \ref{sec:background_dp}.
2) \textcolor{ black}{\emph{This strategy fails to decouple the training and inference runtime, which naturally differ from each other and require distinct optimization techniques.} Moreover, from a parallelism perspective, the training phase necessitates consideration of all model states, whereas the inference phase utilizes only parameters, thus demanding different parallelism approaches. In contrast to a hybrid approach, we argue that it is inefficient to combine different techniques in a hybrid engine. This inefficiency stems not only from the aforementioned communication challenges but also from the inability to leverage more effective, specialized optimization techniques for distinct runtimes. For instance, Megatron could be employed for training, while vLLM could be utilized for generation~\cite{kwon2023efficient}. 3) \emph{Heterogeneous GPU support is difficult under this strategy.} It is due to kernel compatibility and network heterogeneity issues~\cite{ding2021hetseq}, hindering the usage of more affordable and specialized GPU series. Also, obtaining heterogeneous GPUs (e.g., a mixture of A100 and V100) is comparatively easier than homogeneous high-end GPUs in the cluster~\cite{jeon2019analysis,weng2022mlaas}.}

\subsection{Key Challenges}
To overcome these limitations, it is crucial to explore new distributed model placement strategies for efficient RLHF training, which could also save substantial costs for large-scale training prevalent in this field. However, to build an efficient and easy-to-use RLHF training framework, we encounter multiple challenges, including:  

\textbf{Trade-off Between Memory and Communication in Complex System.} 
The design of the Model Placement strategy for the RLHF system presents complexity from two perspectives. Firstly, it requires considering the intricate dependencies among each model in RLHF training. Failure to do so can result in unnecessary data interaction and device idleness as aforementioned in trlX. Secondly, it needs careful trade-offs between memory and communication costs for the collective set of four models in either training or inference mode, as well as for each individual model. This necessitates considering different parallelism strategies or optimization techniques based on the scale of model size and hardware heterogeneity.

\textbf{High Level of Professionalism.} 
The development of an efficient distributed RLHF strategy is not only a demanding task in terms of design and implementation but also poses challenges to users, particularly modeling experts who specialize in machine learning modeling but may lack knowledge in distributed computing. The usage and configuration of distributed RLHF also demand the understanding of the trade-offs involved in the distributed details to some degree. This knowledge gap can potentially hinder the efficient functioning of the distributed RLHF system in practice. Furthermore, due to the dynamic scheduling of GPUs, users are unaware of the hardware specifications during model development, also creating a gap between the model development process and the underlying hardware environment.

\section{Our Framework}
Therefore, we introduce our FlexRLHF Framework for accelerating the RLHF Training. Firstly, besides the Co-located strategy, we propose two model placement strategies aimed at improving the efficiency of RLHF training. These strategies are carefully designed to optimize communication efficiency and utilize memory effectively, catering to different training scenarios that involve different model scales and hardware heterogeneity. Secondly, we present the design of a FlexRLHF Execution Engine abstraction, which provides modeling experts with a simplified programming API by decoupling RLHF modeling from distributed computing. Lastly, we offer a guideline on how to configure the distributed strategies for efficient RLHF training in practice.

\begin{figure*}
\begin{minipage}[t]{0.48\linewidth}
\includegraphics[width=\linewidth]{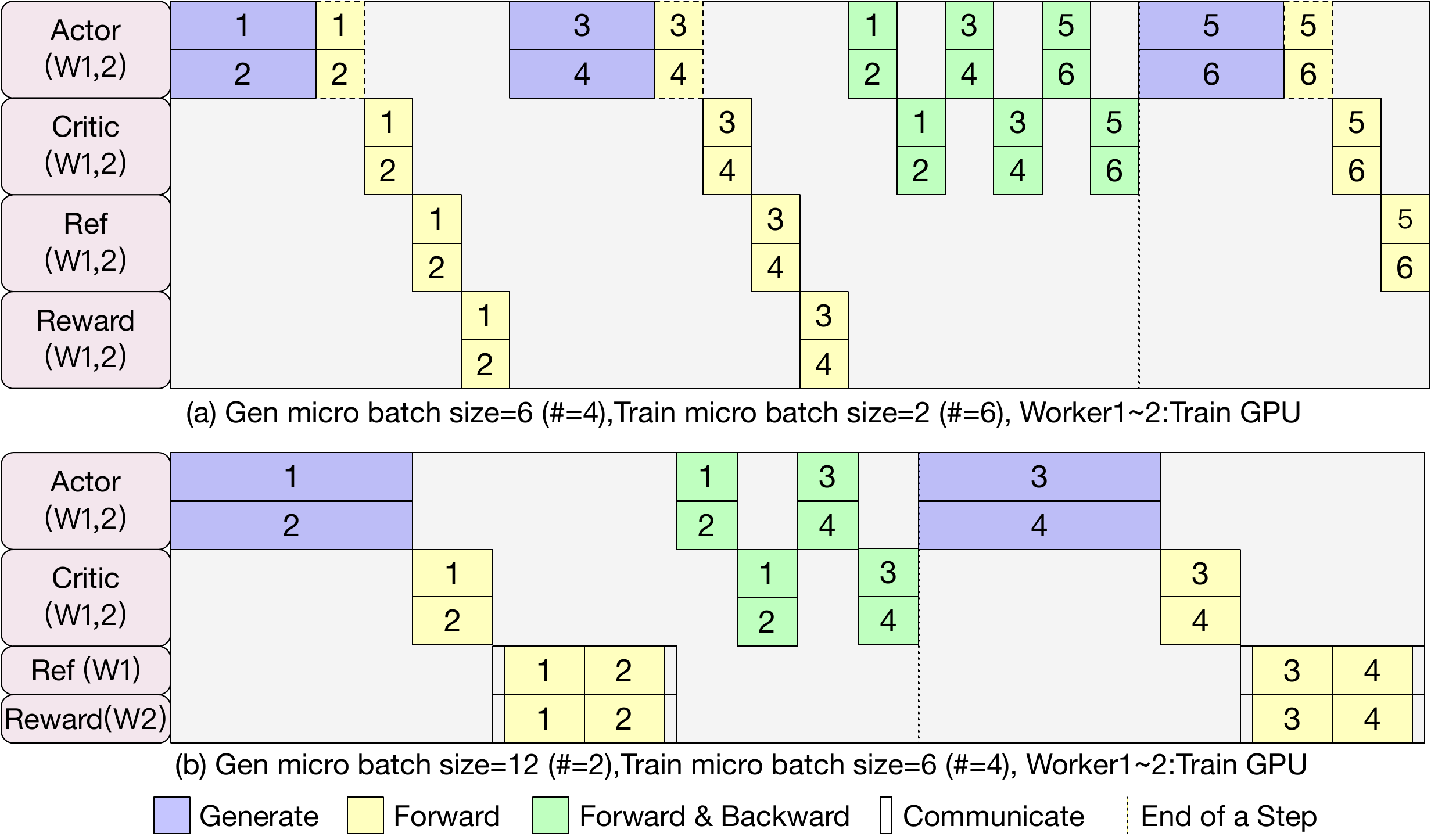}
\caption{The timeline of (a) Co-located strategy vs. (b) Interleaving strategy using two homogeneous GPU devices. We designate ``Worker 1" or ``W1" to represent GPU device 1 and \#=4 means that the number of micro batches is 4. Under the Co-located strategy, both the Ref model and Reward model are deployed on both devices, whereas under the Interleaving strategy, the Ref model and Reward model are allocated to separate groups or devices. The efficiency improvement of our strategy is attributed to two reasons: i) \textbf{Reduced memory redundancy}: In the Co-located strategy (a), four models are allocated on both W$1$ and W$2$, while in the Interleaving strategy (b), the Ref model and Reward model are placed on W$1$ and W$2$ exclusively. This reduces the memory redundancy of parallelism, e.g., low-level ZeRO Parallelism, by reducing participating devices for the Reward model or Ref model from $2$ to $1$. ii) \textbf{Reduced communication cost}: As the Interleaving strategy assigns the Ref model and Reward model to two separate GPU devices, it enables independent and parallel forward computation without communication between two devices.}
\label{fig:pipeline_interleave}
\end{minipage}%
\hfill
\begin{minipage}[t]{0.48\linewidth}
\includegraphics[width=\linewidth]{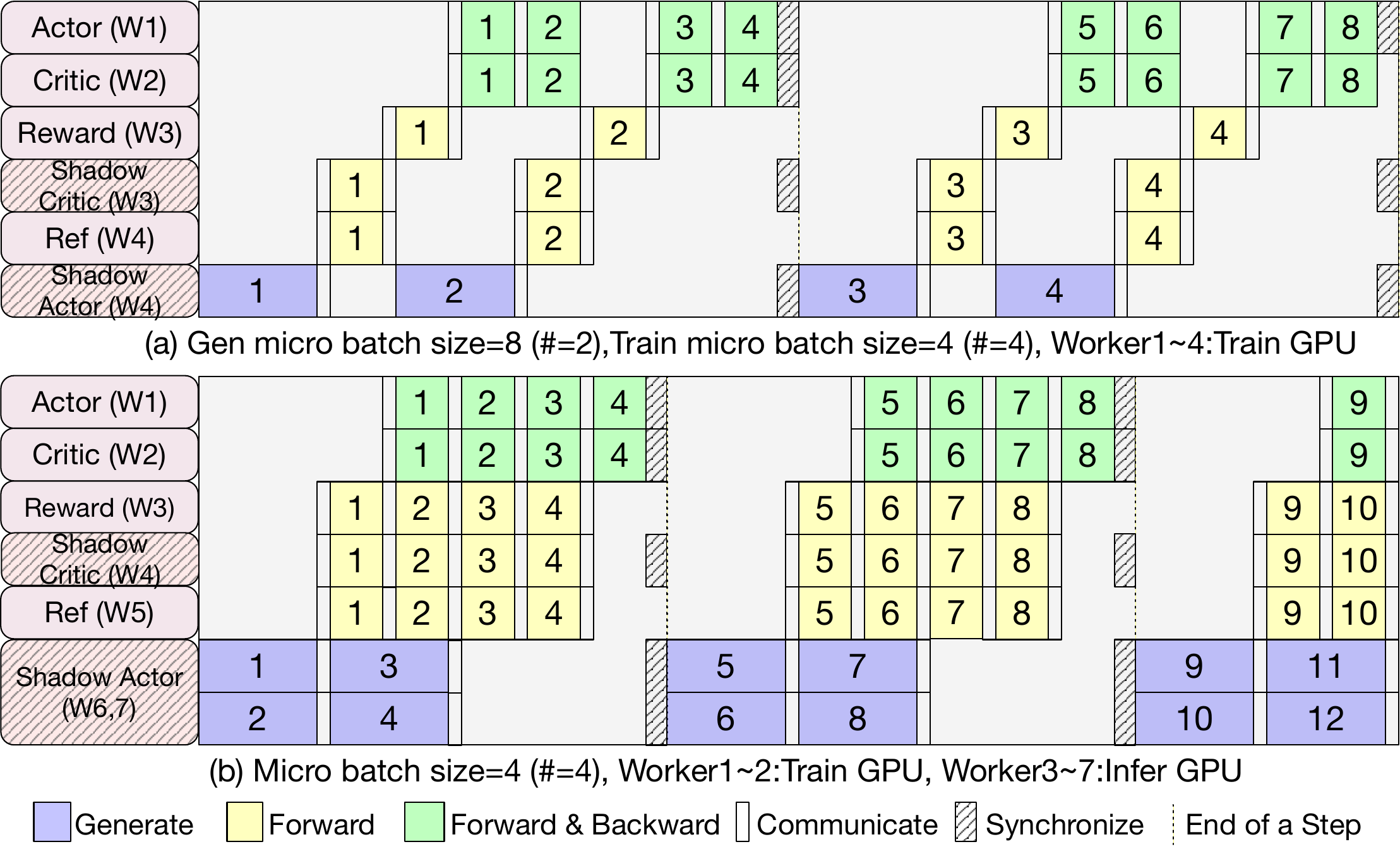}
\caption{The timeline of Disaggregated strategy on homogeneous (a) and heterogeneous GPU devices (b). 
As illustrated in the figure, we set up shadow critic and shadow actor models and placed them on separate GPU devices (as shown in figure (a) placed on W3 and W4, or in figure (b) placed on W4 and W6,7), thereby decoupling the training and inference runtime in the RLHF pipeline. Additionally, in the heterogeneous Disaggregated strategy in (b), these two stages are designated to specialized devices for training or inference purposes. As discussed in Section \ref{sec:sep}, our Disaggregated strategy could benefit from targeted optimization for each stage. Also, this strategy does not require waiting for the entire generation process to complete before proceeding with the Forward and Forward \& Backward operation in a pipeline manner as shown in (b). For instance, in the pipeline execution process, the Forward tasks F1 and F2 can commence immediately following the completion of the Generation tasks G1 and G2, without being blocked by subsequent Generation tasks G3 and G4.}
\label{fig:pipline_split}
\end{minipage}%
\end{figure*}

\alglanguage{pseudocode}
\begin{algorithm}[!ht]
\caption{Interleaving Strategy}
\label{algo:algo1}
\textbf{Input}: $n$ devices; Actor, Critic, Reward, and Ref models

\begin{algorithmic}[1]
\State{ ExpBuffer $\gets$ $\lbrace$$\rbrace$}
\While{not coverage}
\For{RolloutNum in RolloutNums}
\State{ Sampling \emph{Query} from \emph{Training Dataset}}
\State{ \emph{Response} $\gets$ Actor.generate(\emph{Query})}
\State{ \emph{ActorOutput} $\gets$ Actor.forward(\emph{Query}, \emph{Response})}
\State{ \emph{CriticOutput} $\gets$ Critic.forward(\emph{Query}, \emph{Response})}
\State{\textbf{Perform \textit{AllGather} between Reward and Ref models to collect all (\emph{Query}, \emph{Response})}} 
\If{Model in [Ref, Reward]} \Comment{{\textcolor{blue}{Run in parallel and \emph{Output} could be either \emph{RefOutput} or \emph{RewardOutput} depending on current device}}}

\State{ \emph{Output} $\gets$ Model.forward(\emph{Query}, \emph{Response})} 
\EndIf
\State{\textbf{Perform \textit{AlltoAll} between Reward and Ref models to distribute \emph{Output} to all devices.}}

\State{ExpBuffer.add($\langle$\emph{ActorOutput}, ..., \emph{RewardOutput}$\rangle$)}
\EndFor
\For{ppo\_epoch in ppo\_epochs} 
\For{\emph{TrainBatch} in ExpBuffer} 
    \State{Actor.train(\emph{TrainBatch})}
    \State{Critic.train(\emph{TrainBatch})}
\EndFor
\EndFor
\EndWhile
\end{algorithmic}
\end{algorithm}

\subsection{Model Placement Strategies}\label{sec:placement}
\subsubsection{Interleaving Strategy} 
The main idea of the Interleaving strategy is to partially parallelize the pipeline by allocating the independent models to exclusive group of GPU devices, as well as shrinking the number of computing devices required for each model. We observe that either Reward\&Ref Forward (and Actor\&Critic training models) could operate independently, thereby it is possible to eliminate part of the serial execution. For example, the Ref model and Reward model both take the generated responses from the Actor model as input and output the intermediate results into the experience buffer without synchronization. Also, training Actor and Critic nodes have no dependency since they do not need synchronization during the training phase. This design helps the parallelization of independent models in RLHF training. This lowers the memory redundancy and shrinks the scale of participating devices used by independent models rather than using all devices in the Co-located strategy.

We compare the distributed architecture between the Co-located strategy and our Interleaving strategy using a small number of devices as illustrative examples shown in Fig.~\ref{fig:placement}. Following Algorithm~\ref{algo:algo1}, our strategy could be easily extended to large-scale computing devices. In terms of Model Placement, the Co-located strategy treats the four models as a unit and deploys them on all devices, as depicted in Fig. ~\ref{fig:placement}(a) and Fig. ~\ref{fig:pipeline_interleave}. By contrast, our Interleaving strategy distinctly assigns the Ref model and Reward model to two distinct device groups (GPU 0-3 and 4-7)\footnote{Our interleaving strategy could further place the Actor and Critic model on exclusive devices. We only illustrate the interleaving between Reward and Ref for simplicity.}, as illustrated in Fig. ~\ref{fig:placement}(b). It reduces the number of devices by half used by either the Ref model or the Reward model. It's important to note that responses generated from the Actor are located on GPUs where the Ref model is not placed, since the Actor model is deployed on all GPU devices, as shown in the timeline in Fig. ~\ref{fig:pipeline_interleave}(b). Therefore, it is necessary to perform two extra communication stages: an \textit{AllGather} primitive to gather the generated results from all devices before the forward stage, and then an \textit{AlltoAll} primitive to distribute the results over all devices. However, these two additional communication stages can be considered negligible by overlapping with computation. In other words, computation and communication occupy different cuda streams, allowing computation to begin upon receiving the first batch while the remaining batches can be communicated synchronously with the ongoing computation. We conclude that the Interleaving strategy shortens the time-cost generation duration from two aspects:

Firstly, in terms of memory consumption, our Interleaving strategy reduces the number of participating devices for either the Ref or Reward model, effectively minimizing memory redundancy when applying \textcolor{ black}{DP-based Parallelism, e.g., ZeRO}. This allows our method to leverage the saved memory to increase the total batch sizes, resulting in higher overall throughput. For instance, when employing the Co-located strategy under ZeRO-0 as displayed in Fig. ~\ref{fig:pipeline_interleave}(a), all devices hold replicas of the model states, e.g. of Reward Model, while half of devices need to hold the replicas under our Interleaving strategy as depicted in Fig. ~\ref{fig:pipeline_interleave}(b). 

Secondly, from a communication efficiency perspective, having fewer participating devices leads to reduced communication overhead in both the generation and training stages. Also, it's important to note that while increased levels of ZeRO parallelism reduce memory redundancy in model states, it comes at the cost of increased communication as discussed in Section~\ref{sec:background_network}. In this case, the Interleaving strategy can further leverage high-speed intra-node bandwidth and reduce inter-node communication by colocating the Reward model or Ref model on a single machine, particularly in small to medium training scenarios, e.g. 7B or 13B models.

\alglanguage{pseudocode}
\begin{algorithm}
\caption{Disaggregated Strategy}
\label{algo:algo2}
\textbf{Input}: $n$ training devices, and $m$ inference devices; Actor, Critic, Reward, and Ref models.

\begin{algorithmic}[1]
\State{ExpBuffer $\gets$ $\lbrace$$\rbrace$ }

\While{not coverage}\newline
\Comment{\textcolor{blue}{Running following processes on inference devices concurrently}}

\If{Inference Device} 
\State{\emph{Outputs} $\gets$ $\lbrack \rbrack$}
\For{RolloutNum in RolloutNums}
\State{ Sampling \emph{Query} from \emph{Training Dataset}}
\State{ \emph{Response} $\gets$ ShawdowActor.generate(\emph{Query})}

\State{\textbf{Perform \textit{Point-to-Point} in inference devices to distribute $\langle$\emph{Query}, \emph{Response}$\rangle$ over all inference devices}} 
\If{Model in [ShadowActor, ShadowCritic, Ref, Reward]} 
    \State{ \emph{Output} $\gets$ Model.forward(\emph{Query}, \emph{Response})}
    \Comment{{\textcolor{blue}{Run in parallel and \emph{Output} could be either 
    \emph{ShadowActorOutput}, 
 \emph{ShadowCriticOutput}, \emph{RefOutput}, and \emph{RewardOutput}}}}
    \State{ \emph{Outputs}.add(\emph{Output})}
\EndIf
\State{\textbf{Perform \textit{Point-to-Point} in inference devices to distribute \emph{Outputs} over all devices.}} 
\State{\textbf{\textit{Send} \textit{Outputs} to training devices} }
\EndFor
\EndIf \newline
\Comment{\textcolor{blue}{Running following processes on training devices concurrently}}
\If{Training Device}
\State{\textbf{\textit{Recv Outputs} and push \textit{Outputs} in ExpBuffer from inference devices}}

\For{ppo\_epoch in ppo\_epochs} 

\For{\emph{TrainBatch} in ExpBuffer}
     \State{Actor.train(\emph{TrainBatch})}
    \State{Critic.train(\emph{TrainBatch})}
  \EndFor
  \EndFor
  \EndIf
  \State{\textbf{Perform \textit{Synchronization} from Actor/Critic models to shadow Actor/Critic models periodically.}} 
  \EndWhile
  \end{algorithmic}
\end{algorithm}

\subsubsection{Disaggregated Strategy}\label{sec:sep}
The main idea of our Disaggregated strategy is to physically disaggregate the training and the inference runtime of the RLHF pipeline to shorten the generation duration. Specifically, we add extra actor and critic replicas as shadow models for inference only without a mixture usage of runtime as shown in Fig. ~\ref{fig:placement}(c). These shadow models are dedicated to inference and their parameters are synchronized with the corresponding training models periodically. This division not only allows us to run the training and generation stages simultaneously but also leverages optimization techniques tailored for training or inference. It also could easily utilize the extra heterogeneous devices in clusters, since the high-end homogeneous GPU devices are scarce. Overall, this approach significantly accelerates the generation stage and re-balance the workloads among these stages in the pipeline. 

 Following Algorithm~\ref{algo:algo2}, we classify these devices into two groups: training or inference only. This strategy places training models (Actor and Critic models) and inference models, including the Shadow Actor and Critic models, on separate devices.  Firstly, we create the communication groups for either inference or training models for further data communication. Secondly, as shown in Fig. ~\ref{fig:pipline_split}, extra \textit{point-to-point} communications are required before or after the Forward stage of each prediction model. Additionally, the training Actor and Critic models need to pull the intermediate outputs from the inference devices. \textcolor{black}{These extra communication stages introduce some idle time per device, known as bubble overhead. However, these bubbles can be minimized by overlapping computation and communication stages in a pipeline manner. As illustrated in Fig. ~\ref{fig:pipline_split}(a), the bubbles are nearly hidden through micro-batching using the pipeline technique.} Secondly, it is necessary to synchronize the trained model parameters between the training Actor/Critic models and shadow models after each round of PPO training. The synchronization overhead is inevitable but acceptable\footnote{It is critical to clarify that this process is not asynchronous training, but still synchronous training without compromising model accuracy. It leverages the off-policy property of PPO that the training phase involves sampling from a buffer rather than tightly following the forward stage.}. Although this approach introduces some complexity and overhead, it provides significant benefits from three different perspectives:

1) Our Disaggregated strategy decouples training and inference runtime of the pipeline, enabling targeted optimization techniques for each stage, as discussed in Section \ref{sec:limitation}. 
 \textcolor{ black}{For example, the time-consuming Actor Generation can be accelerated by utilizing specialized framework or parallelism strategies that are well-suited for inference workloads and avoid \textit{AllGather} overhead compared with using a Hybrid Engine. 2) The strategy also facilitates the integration of evolving inference optimizations, like PagedAttention. There will be considerable runtime swapping overheads when using a mixture of training and inference runtime.} 3) The strategy allows scaling out additional heterogeneous devices tailored for training or inference workloads, especially for the time-consuming generation stage. This flexibility in either quantity or type of devices offers significant adaptability to diverse training scenarios in practice.

\subsection{FlexRLHF Execution Engine}

\begin{figure}[ht]
  \centering
  \includegraphics[width=1.0 \linewidth]{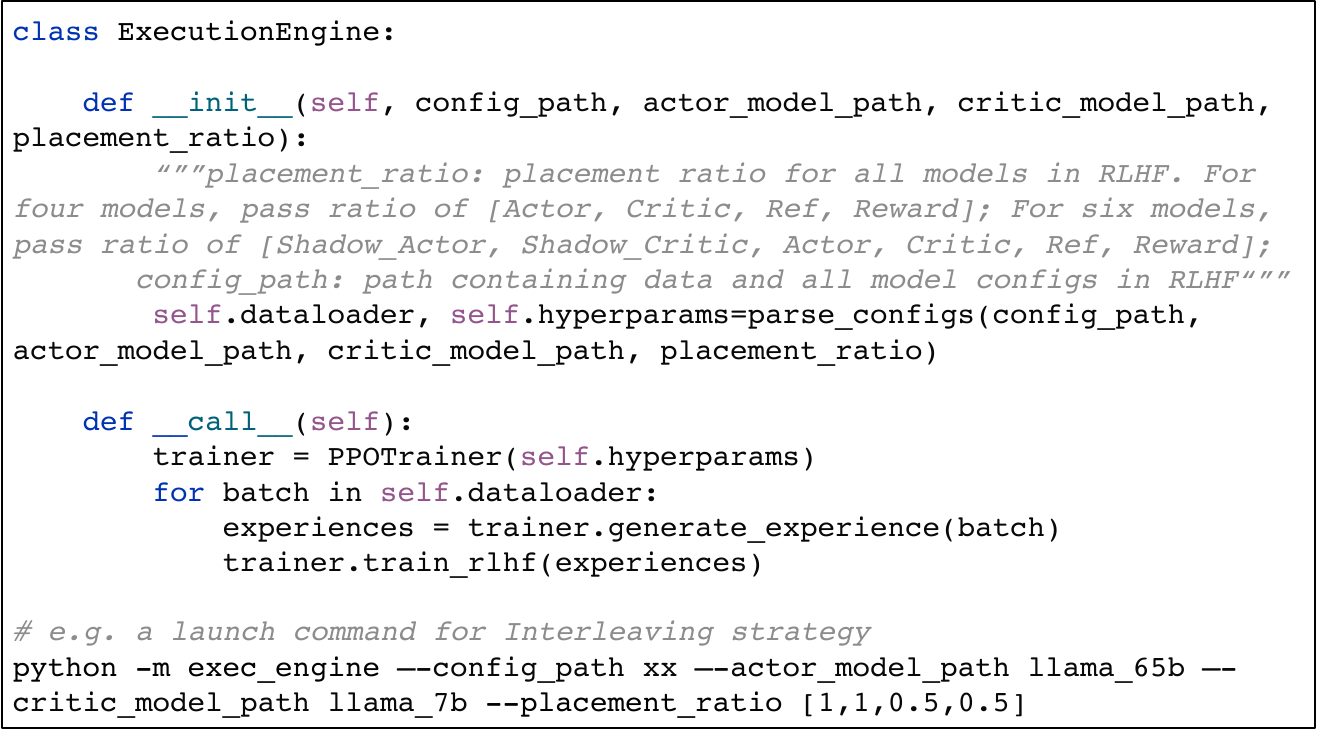}
  \caption{\textcolor{black}{The user interface of FlexRLHF Execution Engine.}}
  \label{fig:code}
\end{figure}

The FlexRLHF Execution Engine is designed to provide a user-friendly interface for modeling experts to easily build their own RLHF training pipeline while hiding the messy details of distributed computing and intermediate data interaction as illustrated in Fig. \ref{fig:code}. The FlexRLHF Execution Engine encapsulates each model involved in the RLHF pipeline considering both AC-Share and AC-NonShare structures. Our modular design allows for the encapsulation of each model and corresponding distributed techniques. These techniques include the model placement strategy, singular model parallelism (such as data parallelism in DeepSpeed and model parallelism in Megatron). We further explain the abstraction in detail.

\textbf{Model Placement Abstraction.} To provide a simple user interface to configure all these strategies, we introduce the concept of Model Placement Ratio, which is a value ranging from 0 to 1 that represents the proportion of devices allocated for a specific model. This offers enhanced adaptability. For instance, the Co-located strategy for a Reward model has a ratio of 1, indicating that the model is deployed on all devices. On the other hand, the Interleaving strategy for the Reward model has a ratio of 0.5, signifying that the model is deployed on half of the devices, as depicted in Fig. \ref{fig:pipeline_interleave}. \textcolor{black}{Take the Reward Model in the model placement ratio representation shown in Fig. \ref{fig:code}, e.g. [1, 1, 0.5, 0.5], as an example. The engine constructs a communication subgroup on the deployed devices of the Reward model for synchronization. It also belongs to a parent communication group that is connected with all models' subgroups for data interaction.} Then our Execution Engine generates the model placement mapping, which places the model to physical devices properly according to the model placement ratio, considering the device capacity and topology, such as intra- and inter-node. This mapping involves detailed rank assignments for each model.
 
\textbf{Hiding Intermediate Data Interaction.} 
 Data interaction caused by the placement strategy is also encapsulated within the Execution Engine. The Interleaving and Disaggregated strategies introduce communication among different communication subgroups during the different stages as discussed in Section \ref{sec:placement}, and the FlexRLHF engine handles these data interaction logics. Each training or forward operation of the models is treated as an individual Operator. The communication group performs an \textit{AllGather} operation to collect the upstream data before executing the Operator. Once the Operator is executed, the resulting data is scattered back downstream.

Overall, this abstraction enables a general interface and backend for different RLHF algorithms beyond PPO, for research or industrial applications. Users can focus on the essence of RLHF modeling without being burdened by the intricacies of the underlying details.

\subsection{A Guideline for Efficient Training}
The guideline addresses the following question: ``\emph{Given $n$ devices and the scale of models, how does one determine the optimal configurations of devices or placement ratio for each model?}" While our proposed placement strategies improve the efficiency and adaptability of RLHF training, there remains a gap in determining the optimal number of devices for each model under different training scenarios where model sizes and the scale of GPU devices vary. 
Distributed training experts could easily employ our model placement ratio abstraction for configuring these strategies. Yet, for modeling experts lacking distributed computing expertise, misconfiguration may downgrade the training efficiency as well as incur frequent occurrences of Out-Of-Memory errors. This difficulty originates from the vast hyperparameter space in the distributed RLHF training pipeline. This often involves coordination between these four or six models and their respective distributed strategies. To address these problems, we introduce the following guidelines to assist modeling experts in the configuration process for more efficient distributed training.

Firstly, in scenarios where computing resources are limited, it is recommended to employ the Interleaving strategy and the Co-located strategy. The interleaving strategy is suitable for cases when either the Ref or Reward model can be accommodated within the memory limits of a single node, including multiple GPU devices. In other cases, the Co-located strategy is advised. This is because inter-node communication is more costly for an Interleaving Strategy than the Co-located strategy, and the Disaggregated strategy requires more memory. However, the usage of the Co-located strategy should be infrequent when computing resources are enough since the Disaggregated strategy becomes a more favorable choice in configurations involving more than two nodes. Furthermore, it is advisable to increase the batch size as long as there is spare memory capacity, stopping before 95\% memory capacity. 

Secondly, employing the Disaggregated strategy proves advantageous when computational resources are enough. 1) To expedite the time-intensive generation phase, it is recommended to allocate additional devices to the Shadow Actor model, provided the minimal memory prerequisites of other models have been satisfied. Guided by our experience, we suggest assigning between 30\% to 50\% of the total computational devices exclusively to the Actor model for generation purposes. \textcolor{ black}{2) Utilizing multiple replicas of the Actor model in a DP manner is beneficial, particularly when inter-node communication is inevitable, as it helps mitigate high communication costs. For instance, employing a replica of the Actor model on each node is preferable to distributing the model over both in scenarios involving two nodes. 3) Furthermore, to optimize throughput using pipeline techniques, it is advisable to configure the model placement ratio for each model to achieve a uniform distribution of processing time across the generation, forward, and training stages. This reduces the bubble overhead of idle devices in different micro-batches and enables stages to work on distinct micro-batches concurrently.}


\section{Experiments}
We conduct comparative experiments on both AC-NonShare and AC-Share model structures against the Co-located strategy used by DeepSpeed-Chat and trlX respectively. These experiments encompass variations in model sizes, device scales, and device types, allowing us to assess the effectiveness and adaptability of our framework across different scenarios.

\subsection{Experiment Settings. }
\textbf{LLM Backbones. } We employ Llama~\cite{touvron2023llama} of different parameter sizes as our LLM backbone, including 7B, 13B, 33B, and 65B. For demonstration purposes, we keep consistent model size for each model involved in RLHF pipeline, but our FlexRLHF framework could adapt to training scenarios where each RLHF model may have varying model sizes and other LLM models, such as GPT~\cite{brown2020language}, and Qwen~\cite{yang2024qwen2}.

\textbf{Hyperparameters.}
The batch size for each experiment is set to the maximum value before Out of Memory. Both the number of PPO training epochs and the number of batches per rollout are set to 1. The input sequence length is 256 and LoRA dim is 4. Both minimum and maximum generation lengths are 256. For other model convergence-related parameters, we keep consistent for fairness.

\textbf{Dataset. } We perform RLHF on the default dataset for DeepSpeed-Chat. At the time of writing, ``Dahoas/rm-static'' hosted on HuggingFace is employed for tuning LLM via RLHF.  This is an open-source ChatBot or Assistant style dataset, specifically designed to create a Helpful \& Harmless conversational system~\cite{bai2022training}. 

\textbf{Baseline. } We conduct a comparison between the FlexRLHF framework and the current state-of-the-art methods in RLHF training: DeepSpeed-Chat~\cite{yao2023deepspeed} with Hybrid Engine for the AC-NonShare case and trlX~\cite{havrilla2023trlx} for the AC-Share case. 

\textbf{Evaluations.} We evaluate the effectiveness of different training frameworks by comparing their sample throughput during training. Sample throughput refers to the rate at which samples are processed end-to-end, typically measured in samples per second(\emph{\# samples/sec}). Furthermore, we verify that the model convergence, measured in returns, remains unaffected by our placement strategies, as evidenced in Fig. \ref{fig:returns}, when compared to the DeepSpeed-Chat implementation.

\textbf{Environments.} Our experiments use the following software versions: CUDA 11.7, DeepSpeed 0.11.2, trlX 0.6.0, PyTorch 1.9.2, Megatron 3.0.2 and NCCL 2.14.3. The experimental cluster utilized in our study comprises up to 16 DGX nodes, with each node equipped with 8 Ampere A100 SXM3 80GB GPUs. The GPUs within each node are interconnected using NVLink, providing a high-speed bidirectional bandwidth of up to 600GB/s. In addition, the nodes are connected via 8 InfiniBand (IB) adapters, which support NVIDIA SHARP and provide around 100GB/s inter-node bandwidth.

\subsection{Experiments and Analysis}
We compare the throughput of our framework against DeepSpeed-Chat and trlX respectively. Specifically, in the training of the 7B model, we utilize ZeRO2 for training models and DP for inference models. For the training of more large models, e.g., 13B, 33B, and 65B models, we have to employ ZeRO3 for both stages. \textcolor{ black}{Furthermore, our Disaggregated strategy leverages tensor and data parallelism in Megatron.}
 
\subsubsection{Comparison in AC-NonShare Structure} 
We implement two Interleaving strategies to compare them with the Co-located strategy used in DeepSpeed-Chat. In $\text{Interleaving}_1$, the Ref and Reward models are placed on half of the GPU devices. In $\text{Interleaving}_2$, we interleave not only the Ref and Reward models but also the Actor and Critic models. Based on our experiments as illustrated in Table \ref{tab:Co-located_vs_interleave}, we conclude that the $\text{Interleaving}_1$ strategy is 12\% faster than the Co-located strategy in DeepSpeed-Chat in small training scenarios, such as 7B model. Additionally, the $\text{Interleaving}_2$ strategy is 71\% faster than DeepSpeed-Chat in larger training scenarios, specifically with 65B models when employing ZeRO-3. The speedup ratio further increases when scaling out to more devices. 

\quad i) \textbf{$\text{Interleaving}_1$ vs. Co-located Strategy}. During the training of 7B size models using 1$\times$8 devices, we observe that $\text{Interleaving}_1$ strategy in our FlexRLHF framework accelerates the throughput from 16.12 to 17.35 samples per second, representing a speedup of 7.63\% compared with Co-located strategy in DeepSpeed-Chat. This speedup ratio further improves to 12\% when scaling out from 1$\times$8 to 4$\times$8 devices. The reason behind this improvement is that the adoption of the interleaving strategy effectively halved the memory redundancy of the Ref and Reward models. Consequently, the freed-up GPU memory can be utilized to expand the overall batch size, as explained in Section \ref{sec:placement}. In this case, the Interleaving strategy achieves a maximum batch size of 36, whereas DeepSpeed-Chat's maximum batch size is 24. However, it is important to note that $\text{Interleaving}_1$ does not perform well with larger model scales, such as in the 33B and 65B training scenarios. This is because these models require ZeRO-3 to reduce the memory redundancy to enable training large language models, while there is no memory redundancy of model states in this case.

ii) \textbf{$\text{Interleaving}_2$ vs. Co-located Strategy.} In contrast, the $\text{Interleaving}_2$ strategy compensates for this drawback in medium to large training scenarios. Although the $\text{Interleaving}_2$ strategy only allocates half of the original GPU count for the Actor model, it exhibits significant improvements when combined with the ZeRO-3 mode. As shown in Table \ref{tab:Co-located_vs_interleave}, the $\text{Interleaving}_2$ strategy is respectively 52\% and 71\% faster than DeepSpeed-Chat in the 33B and 65B model training scenarios when using 2$\times$8 devices. The reason behind this improvement is that, under $\text{Interleaving}_2$, the Actor and Reward models are placed exclusively on a single node, thus avoiding frequent inter-node \textit{AllGather} communication primitives in both the prediction and training stages under ZeRO-3. Additionally, the number of communication participants for the Actor/Critic models is reduced by half.

\begin{table}[]
\caption{Comparing Interleaving Strategy with Co-located Strategy in DeepSpeed-Chat. The results in the table indicate the throughput of the model, measured in samples per second (\emph{\# samples/sec}). The best results are highlighted in bold.}\label{tab:Co-located_vs_interleave}
    \centering
        \resizebox{0.42\textwidth}{!}{
        \begin{tabular}{c|c|cccc}
        \midrule \# of & \multirow{2}{*}{Strategy} & \multicolumn{4}{c}{Model Size} \\ 
         GPUs & & 7B & 13B & 33B & 65B \\
        \midrule \multirow{2}{*}{8} & DeepSpeed-Chat & 16.12 & 7.32 & \textbf{0.69} & OOM \\
         & $\text{Interleave}_1$ & \textbf{17.35} & \textbf{7.63} & \textbf{0.69} & OOM \\
        \midrule \multirow{3}{*}{$2 \times 8$} & DeepSpeed-Chat & 32.15 & 15.35 & 0.64 & 0.14 \\
         & $\text{Interleave}_1$ & \textbf{34.70} & \textbf{15.66} & 0.64 & 0.14 \\
         & $\text{Interleave}_2$ & 18.27 & 8.27 & \textbf{0.97} & \textbf{0.24} \\
        \midrule \multirow{2}{*}{$4 \times 8$} & DeepSpeed-Chat & 61.95 & \textbf{28.60} & 0.93 & \textbf{0.20} \\
         & $\text{Interleave}_1$ & \textbf{69.44} & 28.54 & \textbf{0.95} & \textbf{0.20} \\
        \midrule
        \end{tabular}
        }
\end{table}

\textbf{Disaggregated vs Co-located Strategy.}
The experiments for the Disaggregated strategy are specifically conducted in large-scale training scenarios, such as training 33B and 65B models on $4\times8$ and $16\times8$ GPU devices. This is because this strategy is not well-suited for small training scenarios with a limited number of GPUs and smaller models. The reason is that the Disaggregated strategy involves adding shadow Actor and Critic models exclusively for the prediction stage, which results in increased memory consumption.

However, the benefits of the Disaggregated strategy are significant, as demonstrated in Table \ref{tab:Co-located_vs_Disaggregated}. Our Disaggregated strategy outperforms the Co-located strategy used in DeepSpeed-Chat by factors ranging from 4$\times$ to 11$\times$. For instance, in 33B model training using 4$\times$8 GPU devices, the throughput increases from 0.93 to 4.76 samples per second, resulting in a speedup ratio of 4$\times$. The speedup ratio further increases to 7$\times$ when the model size is increased to 65B. Additionally, the Disaggregated strategy achieves even higher speedup when scaling out to $16\times8$ GPU devices, reaching a speedup ratio of 11$\times$ compared to the baseline using DeepSpeed-Chat. The contribution comes from that the Disaggregated strategy decouples the training and inference runtime of the Actor model, which is not possible for the Co-located strategy. This decoupling allows the usage of intra-node tensor parallelism for the generation stages as discussed in section \ref{sec:placement}. For example, in our training of 65B model using $16\times8$ GPUs, $3\times8$ devices are exclusively allocated for the Shadow Actor model. In this case, we apply tensor parallelism within the same node while replicating it across three machines in a DP manner. \textcolor{ black}{This approach significantly reduces inter-node communications compared to frequent inter-node \textit{AllGather} and \textit{Scatter} communications in ZeRO-3 and resharding in Hybrid Engine.}

\begin{table}[]
    \caption{Homogeneous Disaggregated strategy vs. DeepSpeed-Chat.}
    \centering
    \resizebox{0.32\textwidth}{!}{
    \begin{tabular}{c|c|cc}
\midrule \# of & \multirow{2}{*}{Strategy} & \multicolumn{2}{c}{Model Size} \\ 
 GPUs &  & 33B & 65B \\
\midrule \multirow{2}{*}{$4 \times 8$} & DeepSpeed-Chat & 0.93  & 0.20 \\
 & Disaggregated  & \textbf{4.76} & \textbf{1.75} \\
\midrule \multirow{2}{*}{$16 \times 8$} & DeepSpeed-Chat  & 2.11 & 0.55 \\
 & Disaggregated & \textbf{10.07} & \textbf{6.80}  \\
\midrule
\end{tabular}}

    \label{tab:Co-located_vs_Disaggregated}
\end{table}

\begin{table}[]
    \caption{Heterogeneous Disaggregated strategy vs. DeepSpeed-Chat. We incorporate an additional $2 
    \times 4$ V100 GPUs for inference models to speed up the generation stage.}
    \centering
    \resizebox{0.44\textwidth}{!}{
    \begin{tabular}{c|c|cc}
\midrule \# of & \multirow{2}{*}{Strategy} & \multicolumn{1}{c}{Model Size} \\ 
 GPUs &  & 33B  \\
\midrule \multirow{2}{*}{$1 \times 8 \quad \text{A100}$ } & \multirow{2}{*}{DeepSpeed-Chat} & \multirow{2}{*}{0.69}   \\
 &  &  \\
\midrule
$1 \times 8 \quad \text{A100}$ & \multirow{2}{*}{Heterogeneous Disaggregated} & \multirow{2}{*}{\textbf{0.79}}  \\
 + $2 \times 4  \quad \text{V100}$ &   &  &  \\
\midrule
\end{tabular}}

    \label{tab:Co-located_vs_heterogeneous_Disaggregated}
\end{table}

\textbf{Heterogeneous Disaggregated vs Co-located Strategy.}

We also evaluate the Heterogeneous Disaggregated strategy in a heterogeneous GPU cluster. The cluster consists of 8 A100 GPUs and an additional 8 spare V100 GPUs with 32GB memory. These heterogeneous V100 GPUs can not be easily utilized under the Co-located strategy. We specifically conduct experiments using large models, such as the 33B model. Under our heterogeneous Disaggregated strategy, the Ref and Reward models are placed on a separate node with 4 V100 GPUs respectively. We enable the intra-node tensor parallelism for both models and allocate two A100 GPUs to two Shadow Actor replicas exclusively. This configuration aims to accelerate the generation speed and improve the overall training speed.

As shown in Table \ref{tab:Co-located_vs_heterogeneous_Disaggregated}, introducing extra V100 devices under the heterogeneous Disaggregated strategy results in a speedup of 14.49\%. This acceleration increases the throughput from 0.69 to 0.79 samples per second. Although the acceleration may not be considerable due to the computing power gap between A100 and V100, this strategy explores a novel solution that allows RLHF training to utilize extra heterogeneous resources to enhance the overall throughput. It re-balances the inference workloads among different devices by simply adjusting the batch size, without consideration of software and hardware heterogeneity among different GPUs.

\subsubsection{Comparison in AC-Share Structure}
In the AC-Share scenario, we compare our FlexRLHF framework against trlX. We evaluate our implementation of Interleaving strategies in the FlexRLHF framework in terms of throughput under various computing resources. However, before presenting the experimental results, there are three important notes to mention:

\quad i) To align the number of models with trlX for a fair comparison, our Interleaving strategy only considers the Actor, Ref, and Reward models, excluding the Critic model in terms of placement. The reason is that the Actor and Critic models share most of the parameters in AC-Share structure. The modifications required are minimal, with help of the modular design of our model placement abstraction.

\quad ii) We do not include comparative experiments involving the Disaggregated or Heterogeneous Disaggregated strategies because they are primarily effective for models with large parameter sizes, such as the 33B and 65B models. However, trlX can only place the Reward model on a single GPU, requiring the use of offloading techniques~\cite{paszke2019pytorch} to run the 33B and 65B models. However, offloading will significantly reduce the model's prediction speed. Hence, it can be inferred that our Disaggregated strategy offers more advantages in the trlX scenario compared to results in DeepSpeed-Chat.

\quad iii) In the trlX approach, there are two ways to allocate the Reward model: standalone and coexisted. In the standalone mode, the Reward model exclusively occupies a single GPU, while in the coexisted mode, the Actor model is evenly distributed across all GPUs, and the Reward model is placed on one of them. We conduct a comprehensive comparison of these two modes to evaluate their performance.

\begin{table}[]
\caption{Comparing our Interleaving strategy with trlX  on $1$ machine with $8$ GPU devices. Two kinds of trlX implementation are employed.}
    \centering
    \resizebox{0.41\textwidth}{!}{
\begin{tabular}{c|c|cc}
\midrule Gradient & \multirow{2}{*}{Strategy} & \multicolumn{2}{c}{Model Size} \\ 
 Checkpointing &  & 7B & 13B \\ 
\midrule \multirow{3}{*}{ OFF } & trlX (standalone RM) & 6.04 & 2.84 \\
 & trlX (coexisted RM) & 6.20 & 2.20 \\
 & Interleaving & \textbf{11.17} & \textbf{2.96} \\
\midrule \multirow{3}{*}{ ON } & trlX (standalone RM) & 5.92 & 2.87 \\
 & trlX (coexisted RM) & 6.21 & 2.22 \\
 & Interleaving & \textbf{27.27} & \textbf{11.76} \\
\midrule
\end{tabular}
}

    \label{tab:Co-located_vs_trlx}
\end{table}

\begin{table}[]
\caption{
Comparing our Interleaving strategy with trlX on varying numbers of machines, each with $8$ GPU devices.}
    \centering
    \resizebox{0.35\textwidth}{!}{
\begin{tabular}{c|c|cc}
\midrule \# of & \multirow{2}{*}{Strategy} & \multicolumn{2}{c}{Model Size} \\ 
 GPUs &  & 7B & 13B \\
\midrule \multirow{2}{*}{$1 \times 8$} & trlX (coexisted RM) & 6.21 & 2.22 \\
 & Interleaving & \textbf{27.27}  & \textbf{11.76} \\
\midrule \multirow{2}{*}{$2 \times 8$} & trlX (coexisted RM) & 7.76 & 3.23 \\
 & Interleaving & \textbf{54.71} & \textbf{24.63} \\
\midrule \multirow{2}{*}{$4 \times 8$} & trlX (coexisted RM) & 8.87 & 4.24 \\
 & Interleaving & \textbf{109.18} & \textbf{46.16} \\
\midrule
\end{tabular}
}

    \label{tab:Co-located_vs_interleave_vs_trlx}
\end{table}


\begin{figure*}[t]
    \centering
        \begin{minipage}[t]{0.23\linewidth}
        \centering
        \includegraphics[width=0.9\textwidth]{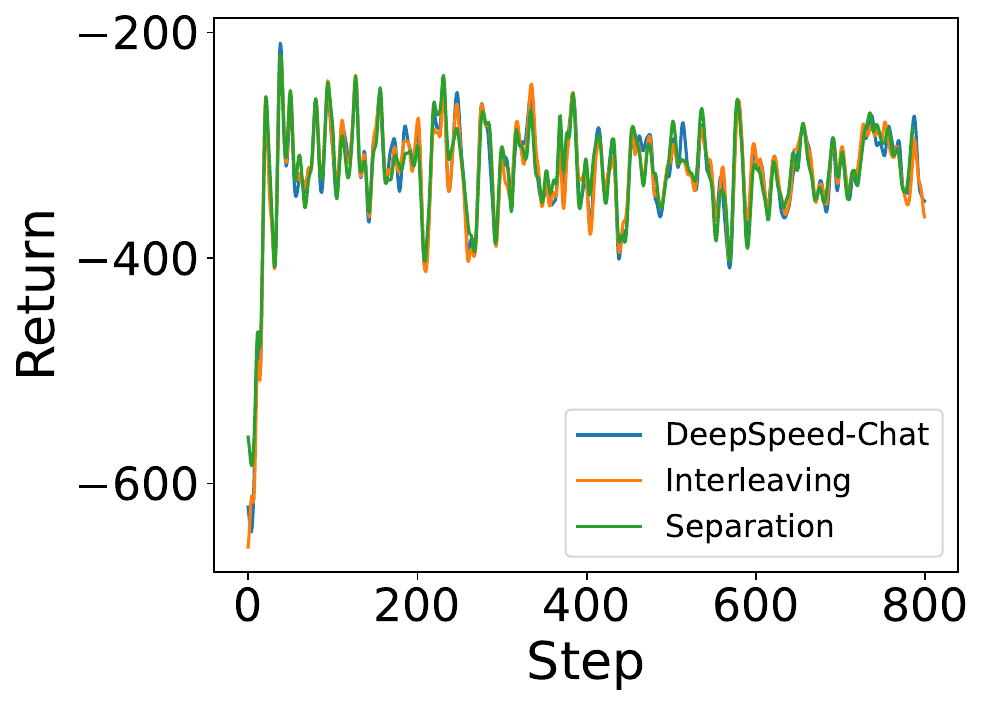}
        \caption{Training convergence between our proposed strategies and DeepSpeed-Chat.}
        \label{fig:returns}
    \end{minipage}
    \quad
    \begin{minipage}[t]{0.23\linewidth}
        \centering
        \includegraphics[width=0.9\textwidth]{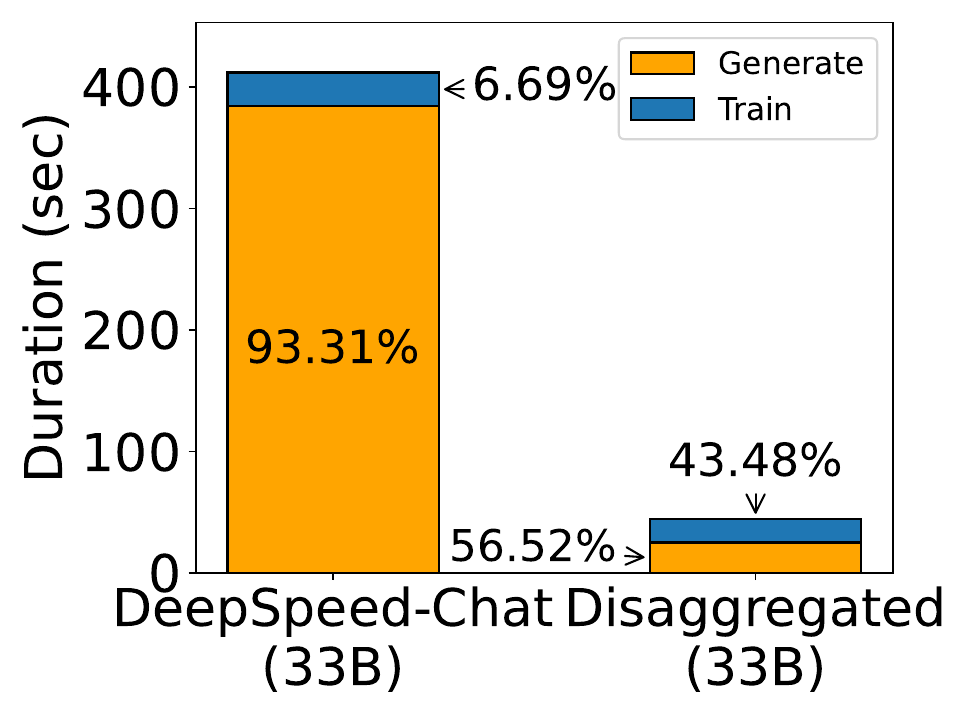}
        \caption{Time cost between Disaggregated strategy and DeepSpeed-Chat in generation and training.}
        \label{fig:time-cost-deepspeed}
    \end{minipage}
    \quad
    \begin{minipage}[t]{0.23\linewidth}
        \centering
        \includegraphics[width=0.9\textwidth]{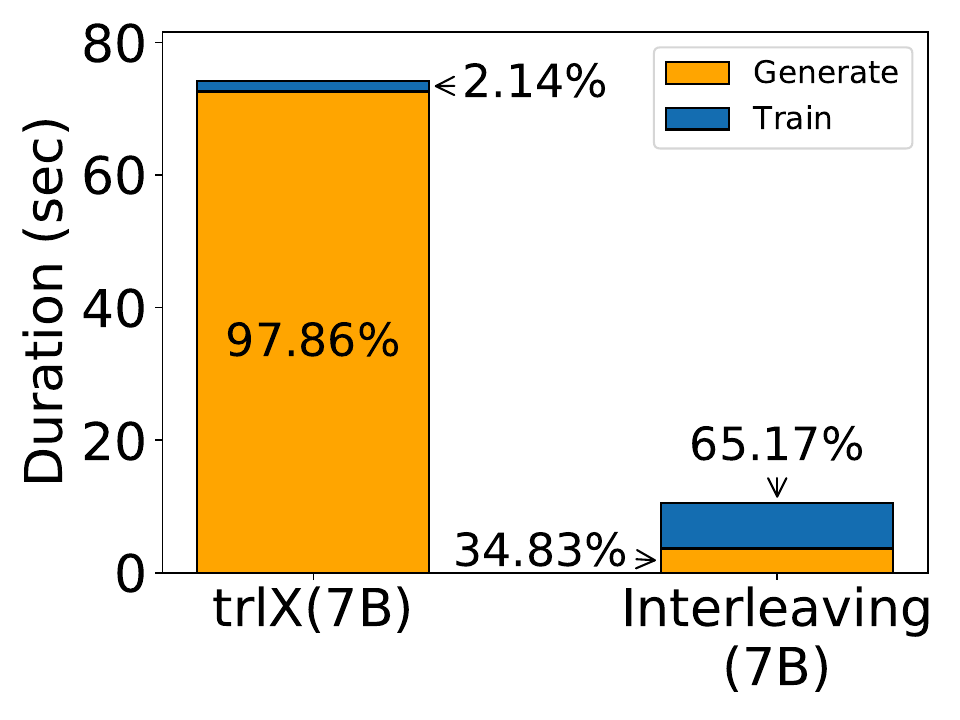}
        \caption{Time cost between Interleaving strategy and trlX in generation and training.}
        \label{fig:time-cost-trlx}
    \end{minipage}
    \quad
    \begin{minipage}[t]{0.23\linewidth}
        \centering
        \includegraphics[width=0.9\textwidth]{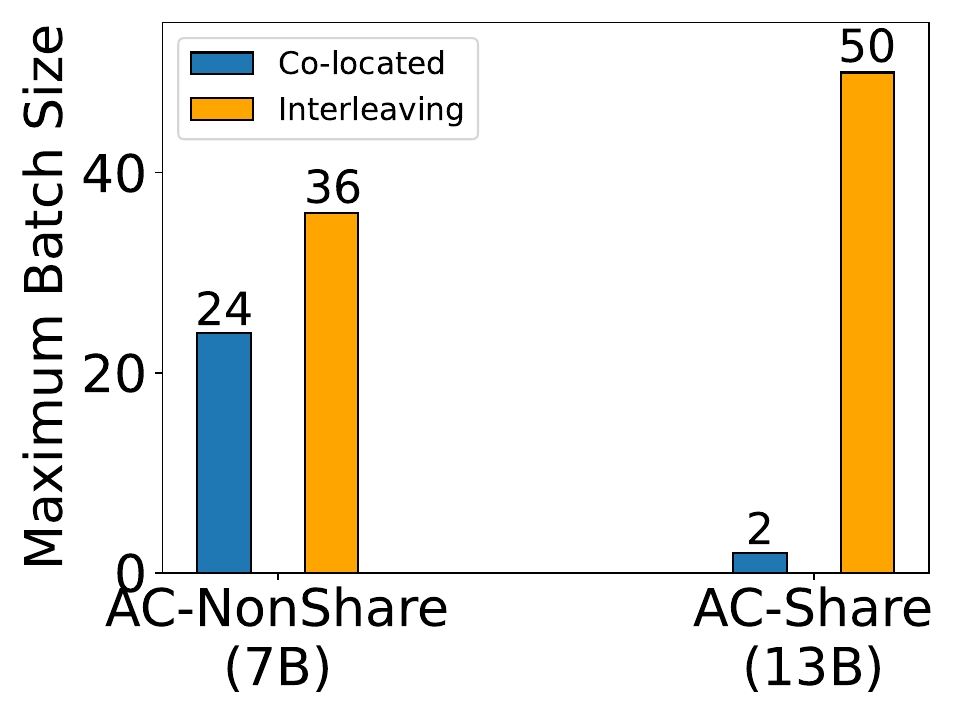}
        \caption{Comparison of maximum available batch size in AC-NonShare and AC-Share scenarios.}
        \label{fig:batch_size}
    \end{minipage}
\end{figure*}


\textbf{Comparison under Gradient Checkpointing.} As shown in Table \ref{tab:Co-located_vs_trlx}, our Interleaving strategy without gradient checkpointing accelerated the throughput from 6.04 to 11.17 samples per second, resulting in a speedup of 84.93\% for training the 6B models. This improvement arises by avoiding the single-device bottleneck of the Reward model in trlX, as discussed in Section \ref{sec:limitation}. Moreover, the remaining memory that is not utilized by the Reward model is wasted, limiting the overall batch size. Also, this design leads to a longer generation duration, leaving other devices idle and resulting in underutilized resources, regardless of whether trlX is in standalone or coexisted mode. It is noteworthy that trlX narrows the performance gap when the Reward model can fully utilize the single device in 13B model training. Additionally, when gradient checkpointing is enabled, the performance gap increases significantly, resulting in a speedup ratio of 360.64\%. This is because gradient checkpointing can reduce memory consumption and in turn, increase the overall training batch size. However, the single-device bottleneck in trlX disables this benefit, hindering the potential performance improvement from gradient checkpointing.

\textbf{Scalability.} 
Our Interleaving strategy demonstrates significant improvements in training speed compared to trlX, which achieves acceleration ranging from $3.4\times$ to $11\times$, when scaling out. As shown in Table \ref{tab:Co-located_vs_interleave_vs_trlx}, our Interleaving strategy increases the throughput from 6.21 to 27.27 samples per second, resulting in a speedup ratio of 3.4$\times$ compared to trlX for the 7B model training on 1$\times$8 GPUs. This speedup ratio further increases to 6.1$\times$ and 11$\times$ when scaling out to 2$\times$8 and 4$\times$8 GPU devices respectively. The training with the 13B models exhibits a similar pattern. It is analyzed that the single-GPU bottleneck in the trlX approach significantly limited the throughput. Furthermore, the interleaving of the Reward model and Ref model in our Interleaving strategy reduces memory redundancy as discussed in Section \ref{sec:placement}, resulting in considerable acceleration.




\subsubsection{Detailed Analysis}
To gain a deeper understanding of the performance enhancements, we conduct further analysis of the time cost for the training and generation stages as well as the maximum batch size available for each strategy. This analysis confirms that our placement strategy significantly improves training efficiency by substantially reducing the time-consuming generation stage and enlarging the maximum batch size.

i) In the AC-NonShare scenario, we experiment with the Llama 33B model on 4$\times$8 GPU devices. Under DeepSpeed-chat's Co-located strategy, the generation stage occupies the majority of the total duration, accounting for 93.31\% in one step, while the training stage only takes up 6.69\% as illustrated in Fig. \ref{fig:time-cost-deepspeed}. However, our Disaggregated strategy reduces the overall duration in one step from around 400 seconds to 30 seconds by significantly reducing the time-consuming generation stage. The percentage of the generation stage is reduced from 93.31\% to 56\%, almost half of the total duration. This reduction is attributed to our Disaggregated strategy, which allows for intra-node tensor parallelism and other inference-optimized techniques for the inference models without \textit{AllGather} parameters for each generation. Additionally, our Interleaving strategy increases 50\% of the maximum batch size available, from 24 to 36 compared with the DeepSpeed-Chat in the 7B model, as displayed in Fig. \ref{fig:batch_size}. 

ii) In the AC-Share scenario, we examine the time expenses for the Llama 7B model on 2$\times$8 GPU devices. As depicted in Fig. \ref{fig:time-cost-trlx}, our approach can reduce the overall duration from around 72 seconds to 10 seconds by largely decreasing the generation stage, as the percentage of generation duration drops from 97.86\% to 34.83\%. In the generation phase, the trlX approach suffers from the aforementioned single-device bottleneck. In contrast, our Interleaving strategy can speed up the generation stage by placing the Reward model on multiple devices and interleaving it with the Ref model. 
For the maximum batch size available, the Interleaving strategy increases the batch size 25$\times$ compared with trlX in the 13B model, rocketing up from 2 to 50, as illustrated in Fig. \ref{fig:batch_size}.

\section{Conclusion}
In this paper, we introduce a novel RLHF training framework that flexibly places multiple LLMs according to the characteristics of RLHF in different training scenarios in terms of model size and device scales. Our two placement strategies, Interleaving and Disaggregated strategies, not only significantly improve training efficiency but also offer agile solutions for various training scenarios. Additionally, our framework is easy to use and provides a simple user interface in practice.

\bibliographystyle{plain}
\bibliography{example_paper.bib}
\end{document}